\documentclass[sigconf,natbib]{acmart}

\AtBeginDocument{%
  \providecommand\BibTeX{{%
    \normalfont B\kern-0.5em{\scshape i\kern-0.25em b}\kern-0.8em\TeX}}}

\setcopyright{acmcopyright}
\copyrightyear{2023}
\acmYear{2023}
\setcopyright{acmlicensed}\acmConference[SIGIR '23]{Proceedings of the 46th International ACM SIGIR Conference on Research and Development in Information Retrieval}{July 23--27, 2023}{Taipei, Taiwan}
\acmBooktitle{Proceedings of the 46th International ACM SIGIR Conference on Research and Development in Information Retrieval (SIGIR '23), July 23--27, 2023, Taipei, Taiwan}
\acmPrice{15.00}
\acmDOI{10.1145/3539618.3592081}
\acmISBN{978-1-4503-9408-6/23/07}

\usepackage{multirow}
\usepackage{hyperref}[colorlinks,linkcolor=blue]
\usepackage{color}

\usepackage{multicol}  
\usepackage{multirow}

\begin{document}

\title{Unsupervised Dialogue Topic Segmentation with \\ Topic-aware Utterance Representation}

\author{Haoyu Gao}
\authornote{Equal Contribution.}
\authornote{Haoyu Gao is also with SIAT, Chinese Academy of Sciences. This work was conducted when Haoyu Gao was interning at Alibaba.}
 \affiliation{%
  \institution{University of Science and Technology of China}
 \state{}
 \country{}
}
\email{haoyugao183@gmail.com}

 \author{Rui Wang}
 \authornotemark[1]
 \affiliation{%
  \institution{Alibaba Group}
 \state{}
 \country{}
}
 \email{wr224079@alibaba-inc.com}

 \author{Ting-En Lin}
 \affiliation{%
  \institution{Alibaba Group}
 \state{}
 \country{}
}
 \email{ting-en.lte@alibaba-inc.com}

 \author{Yuchuan Wu}
 \affiliation{%
  \institution{Alibaba Group}
 \state{}
 \country{}
}
 \email{shengxiu.wyc@alibaba-inc.com}

 \author{Min Yang}
 \authornote{Min Yang and Yongbin Li are corresponding authors.}
 \affiliation{%
  \institution{SIAT, Chinese Academy of Sciences}
 \state{}
 \country{}
}
 \email{min.yang@siat.ac.cn}

 \author{Fei Huang, Yongbin Li}
  \authornotemark[3]
 \affiliation{%
  \institution{Alibaba Group}
 \state{}
 \country{}
}
 \email{shuide.lyb@alibaba-inc.com}

 
\renewcommand{\shortauthors}{Gao and Wang, et al.}

\begin{abstract}
Dialogue Topic Segmentation (DTS) plays an essential role in a variety of dialogue modeling tasks. Previous DTS methods either focus on semantic similarity or dialogue coherence to assess topic similarity for unsupervised dialogue segmentation. However, the topic similarity cannot be fully identified via semantic similarity or dialogue coherence. In addition, the unlabeled dialogue data, which contains useful clues of utterance relationships, remains underexploited. In this paper, we propose a novel unsupervised DTS framework, which learns topic-aware utterance representations from unlabeled dialogue data through neighboring utterance matching and pseudo-segmentation. Extensive experiments on two benchmark datasets (i.e., DialSeg711 and Doc2Dial) demonstrate that our method significantly outperforms the strong baseline methods. For reproducibility, we provide our code and data at: \url{https://github.com/AlibabaResearch/DAMO-ConvAI/tree/main/dial-start}.
\end{abstract}



\maketitle

\section{Introduction}
Dialogue Topic Segmentation (DTS) aims to divide a dialogue into multiple segments, wherein the utterances within each segment are similar in topic. 
DTS is critical in a variety of down-steam dialogue modeling tasks, such as dialogue summarization \cite{chen2020multi, rankae, liu2019automatic, qi2021improving, inan2022structured}, dialogue generation \cite{xu2021discovering, zhang2021modeling, liu2022dial2vec, hu2022unimse}, response prediction \cite{711, lin2022duplex, qian2023empathetic, lin2020discovering, he2022galaxy} and question answering \cite{yoon2018learning, zhang2022slot, dai2022cgodial}.

\begin{figure}[t]
    \centering
    \includegraphics[width=1.1\linewidth]{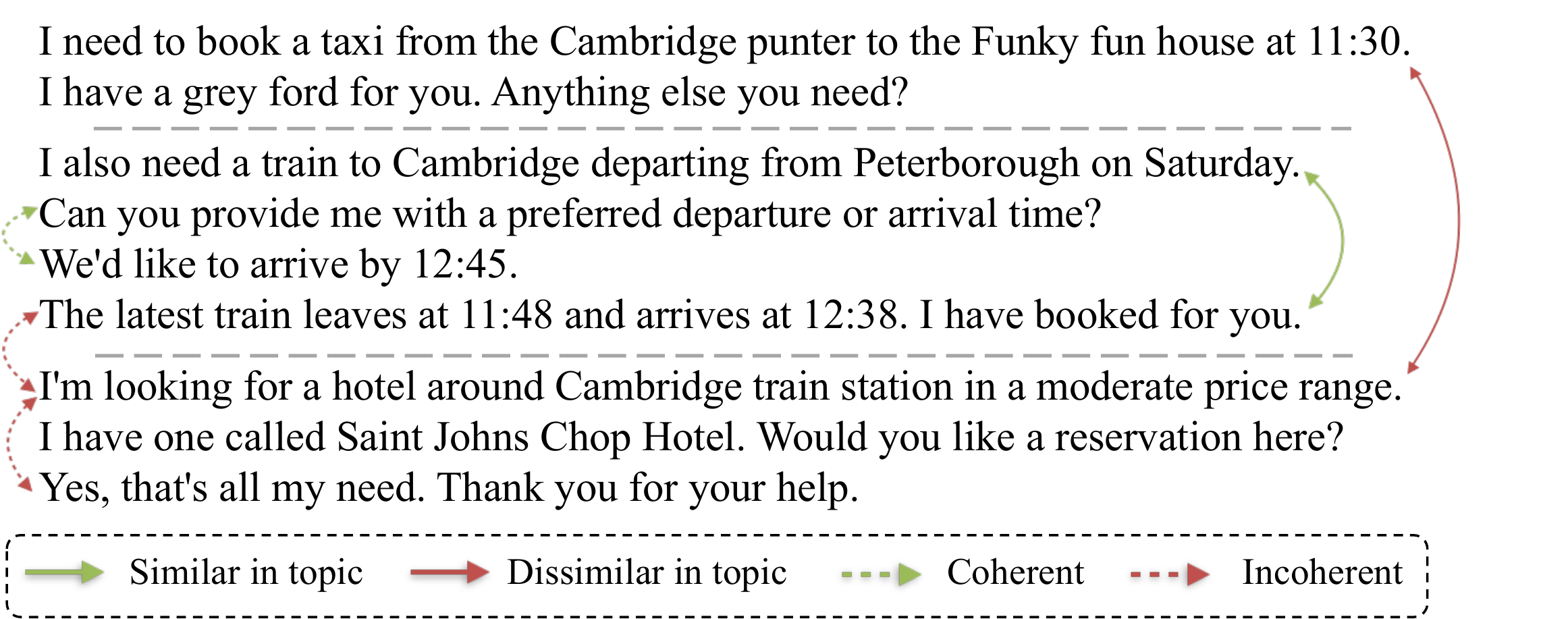}
    \caption{Comparison between topic similarity and dialogue coherence.}
    \label{fig:case}
\end{figure}

Most existing methods for DTS follow an unsupervised paradigm, due to the high cost of collecting accurate DTS annotations to train supervised models \cite{gruenstein2005meeting, xia2022dialogue, zhong2022dialoglm, lo2021transformer} and the variability of annotation instructions across different domains.
These methods generally involve two stages. 
First, various approaches are introduced to assess the topic similarity between the two sides of each potential segment boundary (i.e., an interval between two utterances).
Second, a segmentation algorithm, such as \emph{TextTiling} \cite{texttiling}, is used to identify the segment boundaries. Previous studies usually assess topic similarity through dialogue coherence or semantic similarity computed by surface features, such as lexical overlap \cite{c99, graphseg, TopicTiling}.
In recent years, Song et al. \cite{song2016dialogue} assess semantic similarity using pre-trained word embeddings. The method is later extended to use sentence embeddings from pre-trained language models \cite{711, solbiati2021unsupervised, he2022space, he2022space2}, such as BERT \cite{bert} and SentenceBERT \cite{sentencebert}. 
Xing et al. \cite{csm} further propose Coherence Scoring Model (CSM), which employs utterance-pair coherence to assess topic similarity.

Despite the remarkable progress of previous unsupervised DTS studies, several technical challenges related to modeling topic similarity and utilizing unlabeled dialogue data have not been fully resolved.
First, prior methods typically rely on generic semantic similarity or dialogue coherence to assess topic similarity, but these measures are insufficient to capture it fully.
Specifically, utterances that share the same topic may not be semantically similar, and vice versa.
As illustrated in Figure \ref{fig:case}, dialogue coherence refers to the response relation between an utterance and its preceding context \cite{dziri2019evaluating}, reflecting whether adjacent utterances are linked together. However, two non-adjacent utterances in the same topic segment may be topically similar but not coherent.
Second, unlabeled dialogue data containing useful clues about utterance relationships is beneficial for unsupervised DTS. However, it has not been effectively leveraged in prior works.
In the semantic similarity-based methods, word or sentence embeddings are pre-trained on generic textual corpora and supervised Natural Language Inferring (NLI) datasets \cite{sentencebert}, which are unsuitable for unlabeled dialogue data.
In the coherence-based methods, CSM \cite{csm} learns dialogue coherence from the DailyDialog dataset \cite{li2017dailydialog} without DTS annotations. However, each of these dialogues is about one single topic, and CSM utilizes the dialogue-level topic labels to produce training samples.

To address the above issues, we propose a novel unsupervised DTS framework, called DialSTART (Unsupervised \textbf{Dial}ogue Topic \textbf{S}egmentation with \textbf{T}opic-\textbf{A}ware Utterance \textbf{R}epresen\textbf{T}ation), which learns topic-aware utterance representations from unlabeled dialogue data through neighboring utterance matching (NUM) and pseudo-segmentation. These topic-aware utterance representations are subsequently utilized in combination with the dialogue coherence to perform unsupervised segmentation.
That is, neighboring utterances referring to those appearing together in one dialogue within a certain distance, are more likely to be topically similar.
In unlabeled multiple-topic dialogues, such self-supervision of neighboring utterances is prone to be relatively noisy.
In order to reduce the noise, we further combine the neighboring relation with pseudo-segmentation to produce refined utterance pairs that are assumed to be topically similar or dissimilar.

In practice, we first acquire topic-aware utterance representations via an utterance encoder.
Second, for each utterance interval, we assess the relevance score which reflects the degree to which the two sides are within the same segment.
These relevance scores are utilized by the TextTiling algorithm to perform segmentation. The segmentation results are used for inference or for generating pseudo-segmentation during training.
Third, we generate topically similar and dissimilar pairs for each utterance based on its neighboring utterances and pseudo-segmentation. 
Finally, we fine-tune the utterance encoder to distinguish between topically similar pairs and dissimilar pairs through the marginal ranking loss.
We also design a relevance modeling task to optimize the whole relevance score by distinguishing between real and synthetic fragments.

We conduct experiments on two dialogue topic segmentation datasets. 
The results show that our framework outperforms the state-of-the-art method by 8.03\% on average in terms of Pk error. Further ablative experiments validate the effectiveness of our topic similarity modeling based on the NUM task and pseudo-segmentation for unsupervised DTS. 
Our contributions are threefold:
\begin{enumerate}
    \item We introduce the Neighboring Utterance Matching (NUM) task to learn topic-aware utterance representations, and exploit both topic similarity and dialogue coherence to perform unsupervised dialogue topic segmentation.
    \item We propose to further reduce the self-supervision noise in the NUM task on unlabeled dialogue data by pseudo-segmentation, to obtain topically similar and dissimilar utterance pairs
    \item Experiments have demonstrated that our novel framework outperforms the state-of-the-art method significantly and the effectiveness of topic-aware utterance representation.
\end{enumerate}

\begin{figure*}[t]
    \centering
    \includegraphics[width=\linewidth]{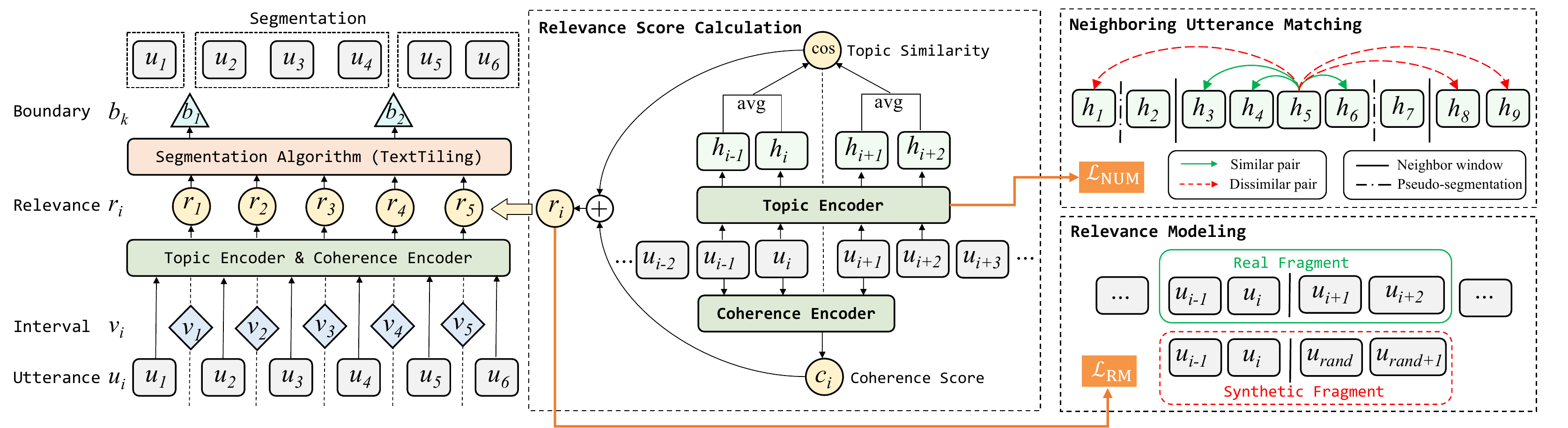}
    \caption{The overview of our model. The left part shows the illustration of segmentation. The middle part represents how to calculate the relevance score based on the output of the topic and coherence encoder. The right part shows the Neighboring Utterance Matching and Relevance Modeling tasks with positive and negative samples.}
    \label{fig:model}
\end{figure*}

\section{Method}
In this section, we introduce our proposed approach. We start with the problem formulation followed by the model overview. Then, we describe topic-aware utterance representation and the training objectives.

\subsection{Problem Formulation}
Dialogue topic segmentation aims to identify segment boundaries in a dialogue.
Formally, given a dialogue $D$ which contains a sequence of $n$ utterances $D = \{u_1, u_2, ..., u_n\}$, there are $n$ - 1 intervals between adjacent utterances, denoted by $V=\{v_1,v_2,...,v_{n-1}\}$. 
A segmentation algorithm predicts segment boundaries as $B =  \{b_1, b_2, ..., b_k\}$, where $k$ denotes the number of boundaries and $b_i$ represents the dialogue is divided at $b_i$-th interval.


Most methods for unsupervised DTS follow a two-stage paradigm. 
First, for the interval $v_i$ which locates between $u_i$ and $u_{i+1}$, a relevance score $r_i$ is computed. The higher the score, the more likely the two sides of the interval belong to the same segment.
Then, given the relevance scores $R = \{r_1, r_2, ..., r_{n-1}\}$, a segmentation algorithm, such as TextTiling \cite{texttiling} or one of its derivatives, is utilized to determine the segmentation boundaries.
Previous methods typically assess the relevance score relying on generic semantic similarity or dialogue coherence. We propose to model that relevance score exploiting both dialogue coherence and \emph{topic similarity} derived from topic-aware utterance representations.

\subsection{Model overview}
As illustrated in Figure 2, our segmentation model consists of a topic encoder, a coherence encoder, and a segmentation algorithm. 
To get a better utterance representation initialization, we choose SimCSE\cite{gao2021simcse} to initialize the topic encoder of our method. SimCSE is a simple but effective contrastive sentence embedding framework. 
We pass $u_i$ into our topic encoder to obtain the topic representations of each utterance: 
\begin{equation}
\mathbf{h_i}=\text{SimCSE}(u_i),
\end{equation}
where $\mathbf{h_i} \in \mathbb{R}^{d_h} $ denotes the pooled output of last layer of SimCSE, $d_h$ is the dimension of hidden state.
Following CSM \cite{csm}, we choose the Next Sentence Prediction (NSP) BERT \cite{bert} as our coherence encoder.
For each interval $v_i$, the coherence encoder calculates the coherence score as
\begin{equation}
c_i=\text{NSP-BERT}([u_{i-1};u_i], u_{i+1}),
\end{equation}
where $u_{i+1}$ is the response and $[u_{i-1};u_i]$ is the concatenated preceding context.
After obtaining the topic representation $\mathbf{h_i}$ and coherence score $c_i$, we calculate the relevance score $r_i$ as
\begin{equation}
r_i=\text{sim}\left(\frac{\mathbf{h}_{i-1}+\mathbf{h}_i}{2}, \frac{\mathbf{h}_{i+1}+\mathbf{h}_{i+2}}{2}\right) + c_i
\end{equation}
where $\text{sim}(\cdot,\cdot)$ is the cosine similarity. The relevance score $r_i$ is then used by the segmentation algorithm to perform segmentation. 

Following most of previous work\cite{song2016dialogue, 711}, we choose TextTiling \cite{texttiling} as our segmentation algorithm. We apply TextTiling on $R=\{r_1,r_2,...,r_{n-1}\}$ to obtain segment boundaries $B$.
\begin{equation}
B = \text{TextTiling}(R).
\end{equation}


To train our encoders, we present two self-supervised task, the neighboring utterance matching task for training the topic encoder and the relevance modeling task for training both encoders.

\subsection{Topic-aware utterance Representation}
In order to train our segmentation model with topic-aware capability, we propose a novel task named Neighboring Utterance Matching (NUM). 
Based on the nature of topic change, we make an assumption that utterance is more likely to be topically similar to its neighboring utterances. To further reduce the noise in unlabeled dialogues, we combine neighboring utterances and pseudo-segmentation to get refined topically similar utterance pairs and dissimilar pairs. Then we take two kinds of pairs as the positive and negative samples for the marginal ranking loss.

First, given an utterance $u_i$ in $D$ , we define its neighboring utterance index set $U_i$ and non-neighboring utterance index set $\overline{U_i}$ as:
\begin{align}
U_i &=\{j \in [1,n] \ | \  w \geq |i-j| \ \land \ j\neq i \}, \\ 
\overline{U_i}&=\{ j \in [1,n] \ |\  w < |i-j| \},
\end{align}
where $w$ is the number of neighboring utterances before and after $u_i$, $n$ refers to the length of the dialogue $D$.

In unlabeled multiple-topic dialogues, the supervision from NUM is prone to be relatively noisy.
In order to reduce the noise, we further combine the neighboring relation with the pseudo-segmentation to produce refined utterance pairs that are assumed to be topically similar or dissimilar. Given $u_i$ and its pseudo segment $segment(i)$, $W_i$ denotes the utterances index inside $segment(i)$ and $\overline{W_i}$ denotes those outside $segment(i)$:
\begin{align}
W_i &= \{j \in [1,n] \ | \ u_j \in \text{segment}(u_i) \ \land j\neq \ i \}, \\ 
\overline{W_i} &= \{ j \in [1,n] \ | \ u_j \notin \text{segment}(u_i)\}.
\end{align}

Based on the neighboring utterances and pseudo segment of $u_i$, we obtain refined topically similar utterances index $P^{+}_i$ and refined topically dissimilar utterances index $P^{-}_i$ for $u_i$ as
\begin{align}
    P^{+}_i = U_i \cap W_i, \quad P^{-}_i = \overline{U_i} \cap \overline{W_i}.
\end{align}
The topic encoder is optimized through the marginal ranking loss:
\begin{equation}
\mathcal{L}_{\rm NUM}(u_i) = \frac{1}{|P^{+}_i|\cdot |P^{-}_i|}\sum_{p^+ \in P^{+}_i}\sum_{p^- \in P^{-}_i}\max(0,\eta + e_{i,p}^-- e_{i,p}^+),
\end{equation}
where $\eta$ is the margin hyper-parameter, $e_{i,p}^-$ and $e_{i,p}^+$ denote for $\text{sim}(h_i,h_{p^-})$ and $\text{sim}(h_i,h_{p^+})$, respectively.


\subsection{Training Objectives}
As illustrated in Figure 2, we utilize two training objectives to enable our segmentation model to capture both dialogue coherence and topic similarity among utterances, combining them for improved segmentation performance. To model topic similarity and obtain better topic representations, we introduce the Neighboring Utterance Matching (NUM) task. Furthermore, based the utterance-pair coherence scoring used in CSM \cite{csm}, we extend it to our relevance modeling (RM) task.
Relevance Modeling (RM) is designed to optimize the topic and coherence encoders to produce high-quality relevance scores. To achieve this, we aim to distinguish between real and synthetic fragments based on the output of the topic and coherence encoders. 
Specifically, for each interval $v_i$, a real fragment consists of the utterances around $v_i$ within a certain distance. In contrast, the synthetic fragment is created by randomly substituting the right-side utterances of $v_i$ while keeping the left-side utterances fixed. 
To simulate various topic transitions, we employ two sampling schemes: 1) sampling utterances only within the same dialogue; 2) randomly sampling utterances from other dialogues.
After obtaining the real and synthetic fragments, we model relevance by ranking real fragments higher than synthetic ones. Formally, the relevance score of real fragment $r_i^+$ centered at $v_i$ is supposed to be higher than the relevance score of the synthetic fragment $r_i^-$. We calculate marginal ranking loss $\mathcal{L}_{\rm RM}$ as follows:
\begin{equation}
\mathcal{L}_{\rm RM}(v_i) = \max(0, \eta + r_i^- - r_i^+).
\end{equation}
Overall, the final training loss is defined as:
\begin{equation}
\mathcal{L} = \frac{1}{N}\sum_{i=1}^{N}\mathcal{L}_{\rm NUM}(u_i) + \frac{1}{M}\sum_{i=1}^{M}\mathcal{L}_{\rm RM}(v_i),
\end{equation}
where $N$ and $M$ is the size of the training data of the NUM task and the RM task, respectively.

\begin{table}[]
\centering
\caption{Experimental results on DialSeg711 and Doc2Dial.}
\label{tab:main_results}
\begin{tabular}{l | cc | cc}
\toprule
\multirow{2}{*}{\textbf{ Method}}  & \multicolumn{2}{c|}{\textbf{DialSeg711}} & \multicolumn{2}{c}{\textbf{Doc2Dial}} \\ 
 & $P_k \downarrow$ & $WD \downarrow$ & $P_k \downarrow$ & $WD \downarrow$ \\
  \midrule
BayesSeg \cite{eisenstein2008bayesian} & 30.97 & 35.60 & 46.65 & 62.13 \\
GraphSeg \cite{graphseg} & 43.74 & 44.76 & 51.54 & 51.59 \\
GreedySeg \cite{711} & 50.95 & 53.85 & 50.66 & 51.56 \\ \midrule
TextTiling (TeT) \cite{texttiling} & 40.44 & 44.63 & 52.02 & 57.42 \\ 
TeT + Embedding \cite{song2016dialogue} & 39.37 & 41.27 & 53.72 & 55.73 \\ 
TeT + CLS \cite{711} & 40.49 & 43.14 & 54.34 & 57.92 \\
TeT + NSP \cite{csm} & 46.84 & 48.50 & 50.79 & 54.86 \\
CSM \cite{csm} & 26.80 & 28.24 & 45.23 & 47.32 \\  \midrule
CSM (unsup) & 24.30 & 26.35 & 45.30 & 49.84 \\
\textbf{Ours} & \textbf{17.86} & \textbf{19.80} & \textbf{38.11} & \textbf{40.72} \\
\bottomrule
\end{tabular}
\end{table}

\section{Experiments}
\subsection{Experimental Setup}
\paragraph{Datasets}
We evaluate our method on two widely used datasets: DialSeg711 and Doc2Dial. 
$\mathbf{DialSeg711}$ \cite{711} is a real-world dataset including 711 English dialogues that combines dialogues from two existing task-oriented dialogue datasets, MultiWOZ \cite{multiwoz} and KVRET \cite{stanford}. This dataset contains 4.9 topic segments per dialogue and 5.6 utterances per topic segment on average.
The $\mathbf{Doc2Dial}$ \cite{feng2020Doc2Dial} dataset comprises more than 4,100 synthetic English conversations grounded in over 450 documents belonging to four domains. This dataset contains 3.7 topic segments per dialogue and 3.5 utterances per topic segment on average.

\paragraph{Evaluation Metrics}
To ensure a fair comparison, we employ two standard metrics, namely $P_k$ error score \cite{pk} and WinDiff (WD) \cite{wd}.
Both $P_k$ and WD are computed by measuring the overlap between the ground-truth segments and the model's predictions within a sliding window of a certain size. 


\paragraph{Implementation details}
We start from the pre-trained checkpoint of the \texttt{sup-simcse-bert-base-uncased} version of SimCSE.
For both DialSeg711 and Doc2Dial, we choose the number of neighboring utterances $w$ as 5 for performance and computational efficiency. 

\subsection{Main Results}
We compare our proposed method with two types of unsupervised baselines: (1) those not using TextTiling including BayesSeg \cite{eisenstein2008bayesian}, GraphSeg \cite{graphseg} and GreedySeg \cite{711}, and (2) those extended from TextTiling such as TeT \cite{texttiling}, TeT+CLS \cite{711}, TeT+Embedding \cite{song2016dialogue}, and  Coherence Scoring Model (CSM) \cite{csm}. Note that CSM uses dialogue coherence instead of semantic similarity, which is different from other unsupervised baselines extended from TextTiling. CSM adopts DailyDialog \cite{li2017dailydialog} as the training data, utilizing the annotations of the dialogue-level topic and utterance-level dialogue act. We also present an improved variant of CSM denoted as CSM (unsup), which does not require topic labels and act labels. Likewise, our framework leverages the unlabeled dialogues for training, while the segmentation annotations are only used for evaluation.


Table \ref{tab:main_results} presents the results of our model and baselines on two datasets. Our model achieves state-of-the-art (SOTA) performance on both evaluation datasets, with varying distances from the previous SOTA. Specifically, we were able to push the performance on DialSeg711 by another about 9\% absolute both $P_k$ error score and WD, achieving 17.86\% $P_k$ error score and 19.80 \% WD. The improvements on Doc2Dial, a more complex and larger dataset, are absolute 7\% on both $P_k$ error score and WD, bringing the SOTA to 38.11\% on $P_k$ error score and 40.72\% on WD. This demonstrates that our model benefits from learning topic similarity and dialogue coherence through effectively exploiting unlabeled dialogues. Furthermore, the CSM(unsup) cannot fully utilize unlabeled data for better improvement, demonstrating the challenge of exploiting unlabeled dialogue. 


\begin{table}[]
\centering
\caption{Ablation study results.}
\label{tab:topic_modeling}
\begin{tabular}{l | cc | cc}
\toprule
\multirow{2}{*}{\textbf{ Method}}   & \multicolumn{2}{c|}{\textbf{DialSeg711}} & \multicolumn{2}{c}{\textbf{Doc2Dial}} \\ 
& $P_k \downarrow$ & $WD \downarrow$ & $P_k \downarrow$ & $WD \downarrow$ \\ \midrule
Ours w/o Topic encoder & 27.60 & 29.84 & 41.34 & 44.36 \\
Ours w/o NUM task & 20.64 & 22.25 & 40.40 & 44.18 \\
Ours w/o Pseudo-seg. & 29.92 & 31.77 & 44.10 & 48.29 \\
\textbf{Ours} & \textbf{17.86} & \textbf{19.80} & \textbf{38.11} & \textbf{40.72} \\
\bottomrule
\end{tabular}
\end{table}

\subsection{Ablation Study}
We investigate the impact of the NUM task and pseudo-segmentation via three different settings: 1) discarding the topic encoder and only utilizing the coherence encoder; 2) keeping the topic encoder but removing the NUM task; 3) keeping the NUM task but removing the pseudo-segmentation. We present our ablation study results in Table \ref{tab:topic_modeling}.
Comparing our approach with the "w/o topic encoder" setting, we observe a significant performance drop on both datasets, indicating that topic similarity is crucial for obtaining good performance, as it prevents local dialogue coherence from dominating the topic segmentation. Removing the NUM task in our method leads to a drop in performance on both datasets to varying degrees, which confirms the effectiveness of our proposed NUM task in encouraging the topic encoder to learn topic-aware utterance representations. Additionally, we found that noise in unlabeled multiple-topic dialogues could mislead the topic encoder's learning, as reflected in the decreased performance of the "w/o pseudo-segmentation" setting.

\section{Conclusion}
In this paper, we propose a novel unsupervised
dialogue topic segmentation framework, which learns topic-aware utterance representations from unlabeled dialogue data through neighboring utterance matching (NUM) and pseudo-segmentation.
Extensive experiments on two benchmark datasets show that our method significantly outperforms previous state-of-the-art by simultaneously utilizing dialogue coherence and topic similarity.

\section*{Acknowledgments}

Min Yang was partially supported by National Key Research and Development Program of China (2022YFF0902100), Shenzhen Science and Technology Innovation Program (KQTD20190929172835662), Shenzhen Basic Research Foundation (JCYJ20210324115614039 and JCYJ20200109113441941). This work was supported by Alibaba Group through Alibaba Innovative Research Program.

\bibliographystyle{ACM-Reference-Format}
\bibliography{sample-base}





\end{document}